\documentclass[sigconf]{acmart}

\renewcommand\footnotetextcopyrightpermission[1]{} 
\usepackage{booktabs} 
\usepackage{etoolbox}
\usepackage{threeparttable}

\newcommand{\ie}{\textit{i}.\textit{e}., }
\newcommand{\eg}{\textit{e}.\textit{g}. }

\begin{document}
\title{You Only Look Twice: Rapid Multi-Scale Object Detection In Satellite Imagery}

\author{Adam Van Etten}
\affiliation{\institution{CosmiQ Works, In-Q-Tel}
}
\email{avanetten@iqt.org}


\renewcommand{\shortauthors}{A. Van Etten}

\begin{abstract}
Detection of small objects in large swaths of imagery is one of the primary problems in satellite imagery analytics. While object detection in ground-based imagery has benefited from research into new deep learning approaches, transitioning such technology to overhead imagery is nontrivial. Among the challenges is the sheer number of pixels and geographic extent per image: a single DigitalGlobe satellite image encompasses $>$ 64 km$^2$ and over 250 million pixels. Another challenge is that objects of interest are minuscule (often only $\sim10$ pixels in extent), which complicates traditional computer vision techniques. 
To address these issues, we propose a pipeline (You Only Look Twice, or YOLT) that evaluates satellite images of arbitrary size 
at a rate of $\geq 0.5$ $\rm{km}^2 / \rm{s}$.
The proposed approach can rapidly detect objects of vastly different scales with relatively little training data over multiple sensors.
We  evaluate large test images at native resolution, and yield scores of $F1 > 0.8$ for vehicle localization.  We further explore resolution and object size requirements by systematically testing the pipeline at decreasing resolution, and conclude that objects only $\sim5$ pixels in size can still be localized with high confidence. Code is available at https://github.com/CosmiQ/yolt
\end{abstract}

%
%



%


\maketitle

\section{Introduction}\label{sec_intro}

Computer vision techniques have made great strides in the past few years since the introduction of convolutional neural networks \cite{alexnet} in the ImageNet \cite{imagenet} competition. The availability of large, high-quality labelled datasets such as ImageNet \cite{imagenet}, PASCAL VOC \cite{pascal_voc} and MS COCO \cite{ms_coco} have helped spur a number of impressive advances in rapid object detection that run in near real-time;  three of the best are: Faster R-CNN \cite{faster_rcnn}, SSD \cite{ssd}, and YOLO \cite{yolov1} \cite{yolo9000}. 
Faster R-CNN 
typically ingests $1000\times600$ pixel images, 
whereas SSD uses $300\times300$ or $512\times512$ pixel input images, 
and YOLO runs on either $416\times416$ or $544\times544$ pixel inputs. 
While the performance of all these frameworks is impressive, none can come remotely close to ingesting the $\sim16,000\times16,000$ input sizes typical of satellite imagery.  
Of these three frameworks, YOLO has demonstrated the greatest inference speed and highest score on the PASCAL VOC dataset. The authors also showed that this framework is highly transferrable to new domains by demonstrating 
superior performance to other frameworks (\ie SSD and Faster R-CNN) on the Picasso Dataset \cite{picasso_dataset} and the People-Art Dataset \cite{people-art_dataset}.  Due to the speed, accuracy, and flexibility of YOLO, we accordingly leverage this system as the inspiration for our satellite imagery object detection framework.

\begin{figure}[t]
\begin{center}
   \includegraphics[width=0.95\linewidth]{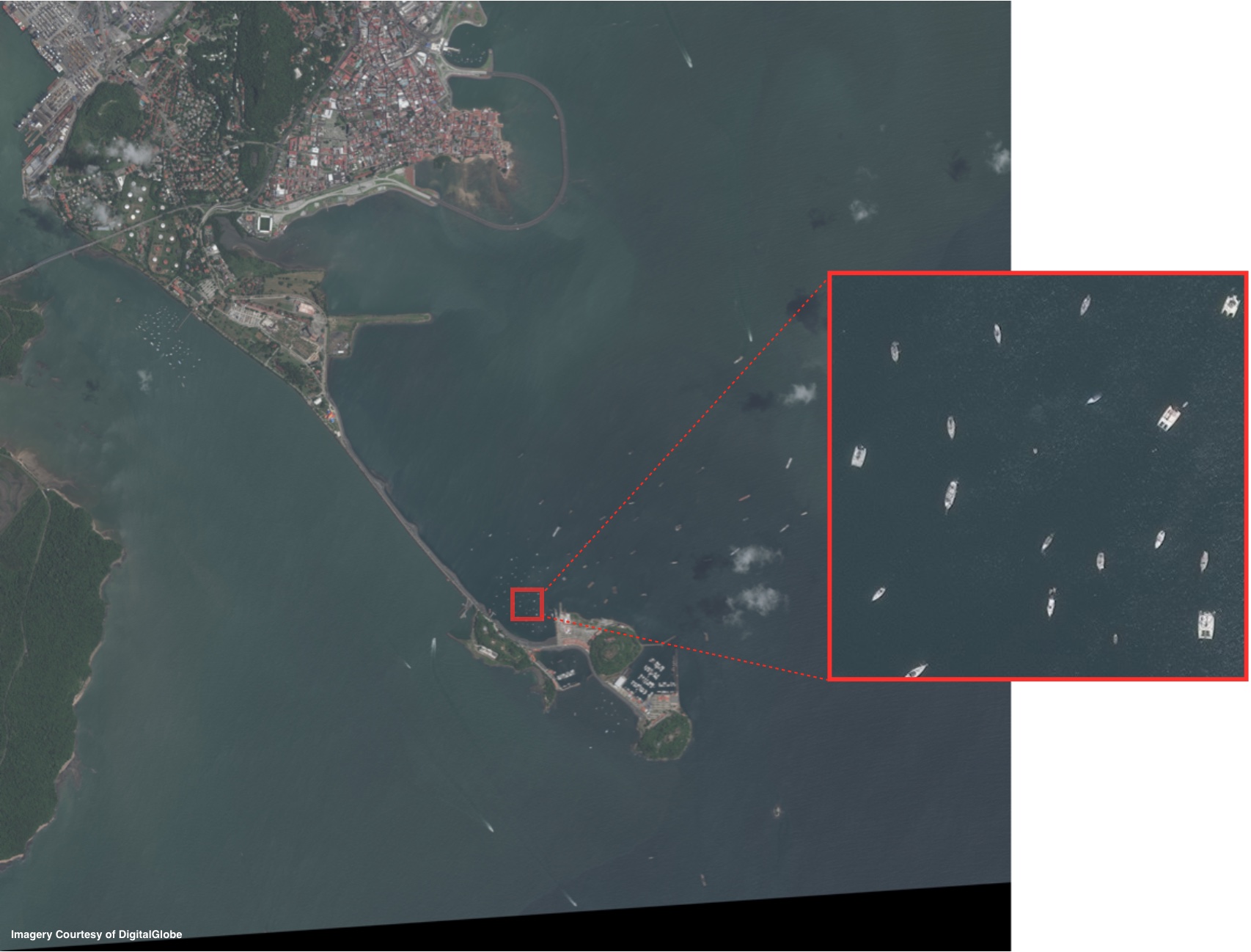}
\end{center}
   \caption{
   DigitalGlobe $8 \times 8$ km ($\sim16,000 \times16,000$ pixels) image at 50 cm GSD near the Panama Canal. One $416 \times 416$ pixel sliding window cutout is shown in red. For an image this size, there are $\sim1500$ unique cutouts.
   }
\label{fig:panama_zooma}
\end{figure}

The application of deep learning methods to traditional object detection pipelines is non-trivial for a variety of reasons. The unique aspects of satellite imagery necessitate algorithmic contributions to address challenges related to the spatial extent of foreground target objects, complete rotation invariance, and a large scale search space. Excluding implementation details, algorithms must adjust for:

\begin{description}
\item[Small spatial extent] In satellite imagery objects of interest are often very small and densely clustered, rather than the large and prominent subjects typical in ImageNet data. In the satellite domain, resolution is typically defined as the ground sample distance (GSD), which describes the physical size of one image pixel.  Commercially available imagery varies from 30 cm GSD for the sharpest DigitalGlobe imagery, to $3-4$ meter GSD for Planet imagery. This means that for small objects such as cars each object will be only $\sim15$ pixels in extent even at the highest resolution.
\item[Complete rotation invariance] Objects viewed from overhead can have any orientation  (\eg ships can have any heading between 0 and 360 degrees, whereas trees in ImageNet data are reliably vertical).
\item[Training example frequency] There is a relative dearth of training data (though efforts such as SpaceNet\footnote{https://aws.amazon.com/public-datasets/spacenet/} are attempting to ameliorate this issue)
\item[Ultra high resolution] Input images are enormous (often hundreds of megapixels), so simply downsampling to the input size required by most algorithms (a few hundred pixels) is not an option (see Figure \ref{fig:panama_zooma}).
\end{description}

The contribution in this work specifically addresses each of these issues separately, while leveraging the relatively constant distance from sensor to object, which is well known and is typically $\sim400$ km. This coupled with the nadir facing sensor results in consistent pixel size of objects. 

Section \ref{sec_related_work} details in further depth the challenges faced by standard algorithms when applied to satellite imagery. The remainder of this work is broken up to describe the proposed contributions as follows.  To address small, dense clusters, Section \ref{sec_arch} describes a new, finer-grained network architecture. Sections \ref{sec_test_proc} and \ref{sec_post_proc} detail our method for splitting, evaluating, and recombining large test images of arbitrary size at native resolution. With regard to rotation invariance and small labelled training dataset sizes, Section \ref{sec_train} describes data augmentation and size requirements.  Finally, the performance of the algorithm is discussed in detail in Section~\ref{sec_results}.

\section{Related Work}\label{sec_related_work}


Deep learning approaches have proven effective for ground-based object detection, though current techniques are often still suboptimal for overhead imagery applications.  
For example, small objects in groups, such as flocks of birds present a challenge ~\cite{yolov1}, caused in part by the multiple downsampling layers of all three convolutional network approaches listed above (YOLO, SDD, Faster-RCNN).  Further, these multiple downsampling layers result in relatively course features  for object differentiation; this poses a problem if objects of interest are only a few pixels in extent.  For example, consider the default YOLO network architecture, which downsamples by a factor of 32 and returns a $13\times13$ prediction grid; this means that object differentiation is problematic if object centroids are separated by less than 32 pixels.  Accordingly we implement a unique network architecture with a denser final prediction grid.  This improves performance by yielding finer grained features to help differentiate between classes.  This finer prediction grid also permits classification of smaller objects and denser clusters.  

Another reason object detection algorithms struggle with satellite imagery is that they have difficulty generalizing objects in new or unusual aspect ratios or configurations~\cite{yolov1}.  Since objects can have arbitrary heading, this limited range of invariance to rotation is troublesome.  Our approach remedies this complication with rotations and augmentation of data. Specifically,  we rotate training images about the unit circle to ensure that the classifier is agnostic to object heading, and also randomly scale the images in HSV (hue-saturation-value) to increase the robustness of the classifier to varying sensors, atmospheric conditions, and lighting conditions.

In advanced object detection techniques  the network sees the entire image at train and test time. While this greatly improves background differentiation since the network encodes contextual (background) information for each object, the memory footprint on typical hardware (NVIDIA Titan X GPUs with 12GB RAM) is infeasible for a 256 megapixel image. 

We also note that the large sizes satellite images preclude simple approaches to some of the problems noted above.  
For example, upsampling the image to ensure that objects of interest are large and dispersed enough for standard architectures is infeasible, since this approach would also increase runtime many-fold.  Similarly, 
running a sliding window classifier across the image to search for objects of interest quickly becomes computationally intractable, since multiple window sizes will be required for each object size.  For perspective, one must evaluate over one million sliding window cutouts if the target is a 10 meter boat in a DigitalGlobe image.  
Our response is to leverage rapid object detection algorithms to evaluate satellite imagery with a combination of local image interpolation on reasonably sized image chips ($\sim 200$ meters) and a multi-scale ensemble of detectors.

\begin{figure}[t]
\begin{center}
   \includegraphics[width=0.95\linewidth]{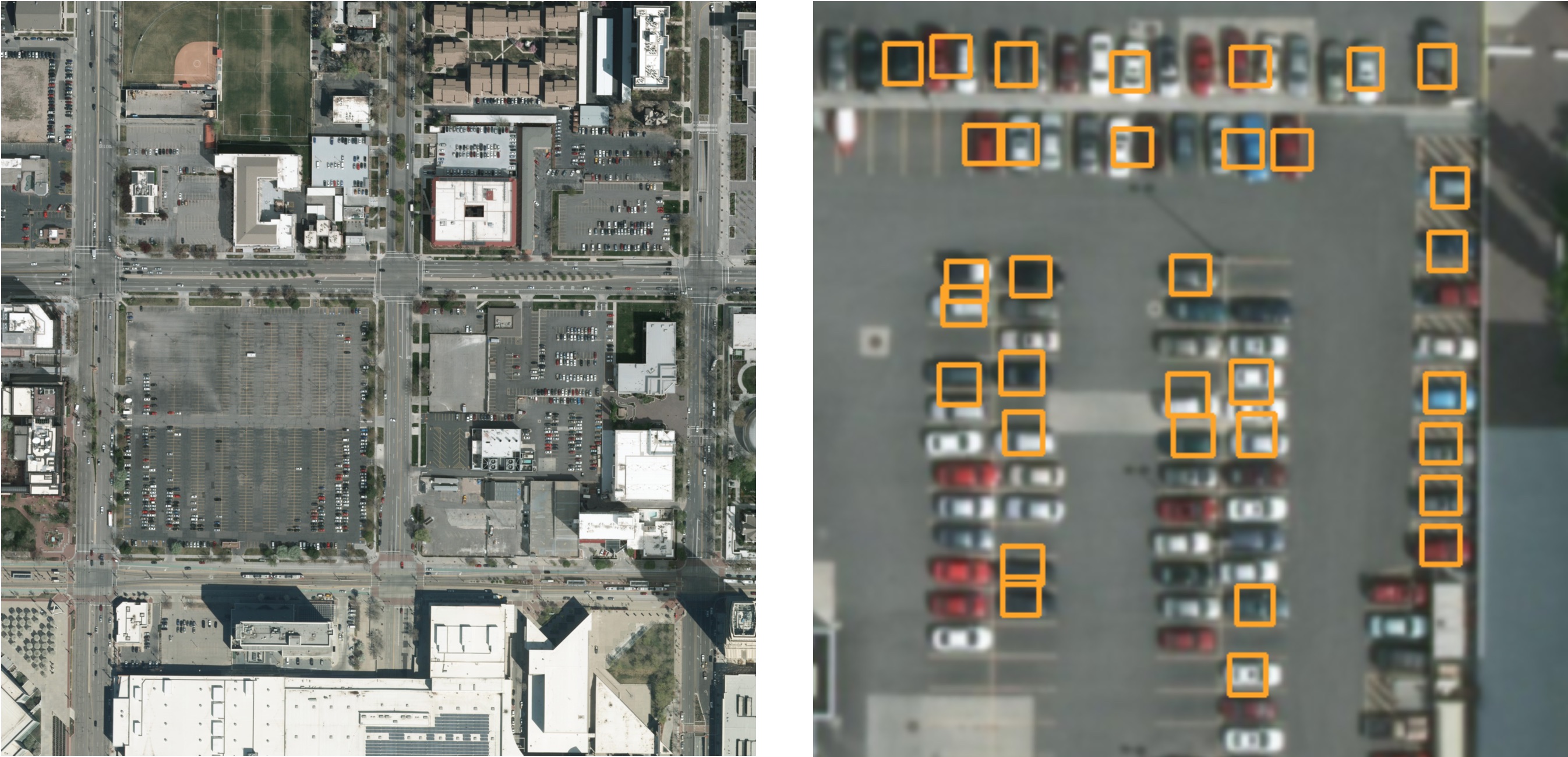}
\end{center}
   \caption{
   Challenges of the standard object detection network architecture when applied to overhead vehicle detection.  Each image uses the same standard YOLO architecture model trained on $416\times416$ pixel cutouts of cars from the COWC dataset.  Left: Model applied to a large $4000\times4000$ pixel test image downsampled to a size of $416\times416$; none of the 1142 cars in this image are detected.  Right: Model applied to a small $416\times416$ pixel cutout; the excessive false negative rate is due to the high density of cars that cannot be differentiated by the $13\times13$ grid.
   }
\label{fig:stock_yolo}
\end{figure}

To demonstrate the challenges of satellite imagery analysis, we train a YOLO model with the standard network architecture ($13\times13$ grid) to recognize cars in $416\times416$ pixel cutouts of the COWC overhead imagery dataset \cite{cowc} (see Section~\ref{sec_train} for further details on this dataset).  Naively evaluating a large test image (see Figure \ref{fig:stock_yolo}) with this network yields a $\sim100\%$ false positive rate, due to the $100\times$ downsampling of the test image.  
Even appropriately sized image chips are problematic (again, see Figure \ref{fig:stock_yolo}), as the standard YOLO network architecture cannot differentiate objects with centroids separated by less than 32 pixels. Therefore even if one restricts attention to a small cutout, performance is often poor in high density regions with the standard architecture.

\section{You Only Look Twice}\label{sec_algo}


In order to address the limitations discussed in Section \ref{sec_related_work}, we implement 
an object detection framework optimized for overhead imagery: 
You Only Look Twice (YOLT).  
We extend the Darknet neural network framework \cite{darknet} and update a number of the C libraries to enable analysis of geospatial imagery and integrate with external python libraries.  
We opt to leverage the flexibility and large user community of python for pre- and post-processing.  Between the updates to the C code and the pre and post-processing code written in python, interested parties need not have any knowledge of C to train, test, or deploy 
YOLT 
models. 

\begin{figure}[t]
\begin{center}
   \includegraphics[width=0.95\linewidth]{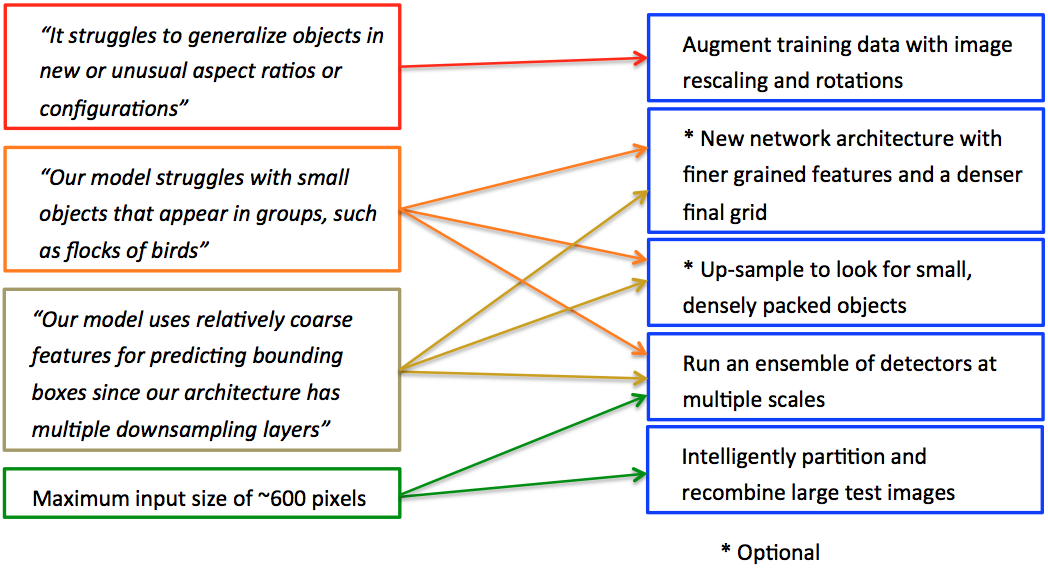}
\end{center}
	\caption{Limitations of the YOLO framework (left column, quotes from \cite{yolov1}), along with YOLT contributions to address these limitations (right column).}
\label{fig:mod_flow}
\end{figure}

\subsection{Network Architecture}\label{sec_arch}


To reduce model coarseness and accurately detect dense objects (such as cars or buildings), we implement a network architecture that uses 22 layers and downsamples by a factor of 16
Thus, a $416\times416$ pixel input image yields a $26\times26$ prediction grid. 
Our architecture is inspired by the 30-layer YOLO network, though this new architecture is optimized for small, densely packed objects.  
The dense grid is unnecessary for diffuse objects such as airports, but crucial for high density scenes such as parking lots (see Figure \ref{fig:stock_yolo}).
To improve the fidelity of small objects, we also include a passthrough layer (described in \cite{yolo9000}, and similar to identity mappings in ResNet \cite{resnet}) that concatenates the final $52\times52$ layer onto the last convolutional layer, allowing the detector access to finer grained features of this expanded feature map.  
 
Each convolutional layer save the last is batch normalized with a leaky rectified linear activation, save the final layer that utilizes a linear activation. The final layer provides predictions of bounding boxes and classes, and has size:
$N_f = N_{\rm{boxes}} \times (N_{\rm{classes}} + 5)$, 
where $N_{\rm{boxes}}$ is the number of boxes per grid (5 by default), and $N_{\rm{classes}}$ is the number of object classes \cite{yolov1}.

\begin{table}[htbp]
\small
\centering
\caption{YOLT Network Architecture} 
\label{tab_SIMRS_arch}
\begin{tabular}{l c c c c}
Layer & Type & 		Filters & 	Size/Stride & 	Output Size  \\
\hline
0 & 	Convolutional &	 	32 & 		3$\times$3 / 1 &		416$\times$416$\times$32 \\
1 & 	Maxpool &		&		2$\times$2 / 2 &		208$\times$208$\times$32 \\
2 & 	Convolutional &	 	64 & 		3$\times$3 / 1 &		208$\times$208$\times$ 64 \\
3 & 	Maxpool &		&		2$\times$2 / 2 &		104$\times$104$\times$ 64 \\
4 & 	Convolutional &	 	128 & 	3$\times$3 / 1 &		104$\times$104$\times$128 \\
5 & 	Convolutional &	 	64 & 		1$\times$1 / 1 &		104$\times$104$\times$64 \\
6 & 	Convolutional &	 	128 & 	3$\times$3 / 1 &		104$\times$104$\times$128 \\
7 & 	Maxpool &		&		2$\times$2 / 2 &		52$\times$52$\times$64 \\
8 & 	Convolutional &	 	256 & 	3$\times$3 / 1 &		52$\times$ 52$\times$256 \\
9 & 	Convolutional &	 	128 & 	1$\times$1 / 1 &		52$\times$ 52$\times$128 \\
10 & 	Convolutional &	 	256 & 	3$\times$3 / 1 &		52$\times$ 52$\times$256 \\
11 & 	Maxpool &		&		2$\times$2 / 2 &		26$\times$ 26$\times$256 \\
12 & 	Convolutional &	 	512 & 	3$\times$3 / 1 &		26$\times$ 26$\times$512 \\
13 & 	Convolutional &	 	256 & 	1$\times$1 / 1 &		26$\times$ 26$\times$256 \\
14 & 	Convolutional &	 	512 & 	3$\times$3 / 1 &		26$\times$ 26$\times$512 \\
15 & 	Convolutional &	 	256 & 	1$\times$1 / 1 &		26$\times$ 26$\times$256 \\
16 & 	Convolutional &	 	512 & 	3$\times$3 / 1 &		26$\times$ 26$\times$512 \\
17 & 	Convolutional &	 	1024 & 	3$\times$3 / 1 &		26$\times$ 26$\times$1024 \\
18 & 	Convolutional &	 	1024 & 	3$\times$3 / 1 &		26$\times$ 26$\times$1024 \\
19 & Passthrough &		&		10 $\rightarrow$ 20  & 	26$\times$ 26$\times$1024 \\
20 & 	Convolutional &	 	1024 & 	3$\times$3 / 1 &		26$\times$26$\times$1024 \\
21 & 	Convolutional &	 	$N_f$ & 	1$\times$1 / 1 &		26$\times$26$\times$$N_f$ \\
\end{tabular}
\end{table}

\subsection{Test Procedure}\label{sec_test_proc}

\begin{figure}[t]
\begin{center}
\includegraphics[width=0.95\linewidth]{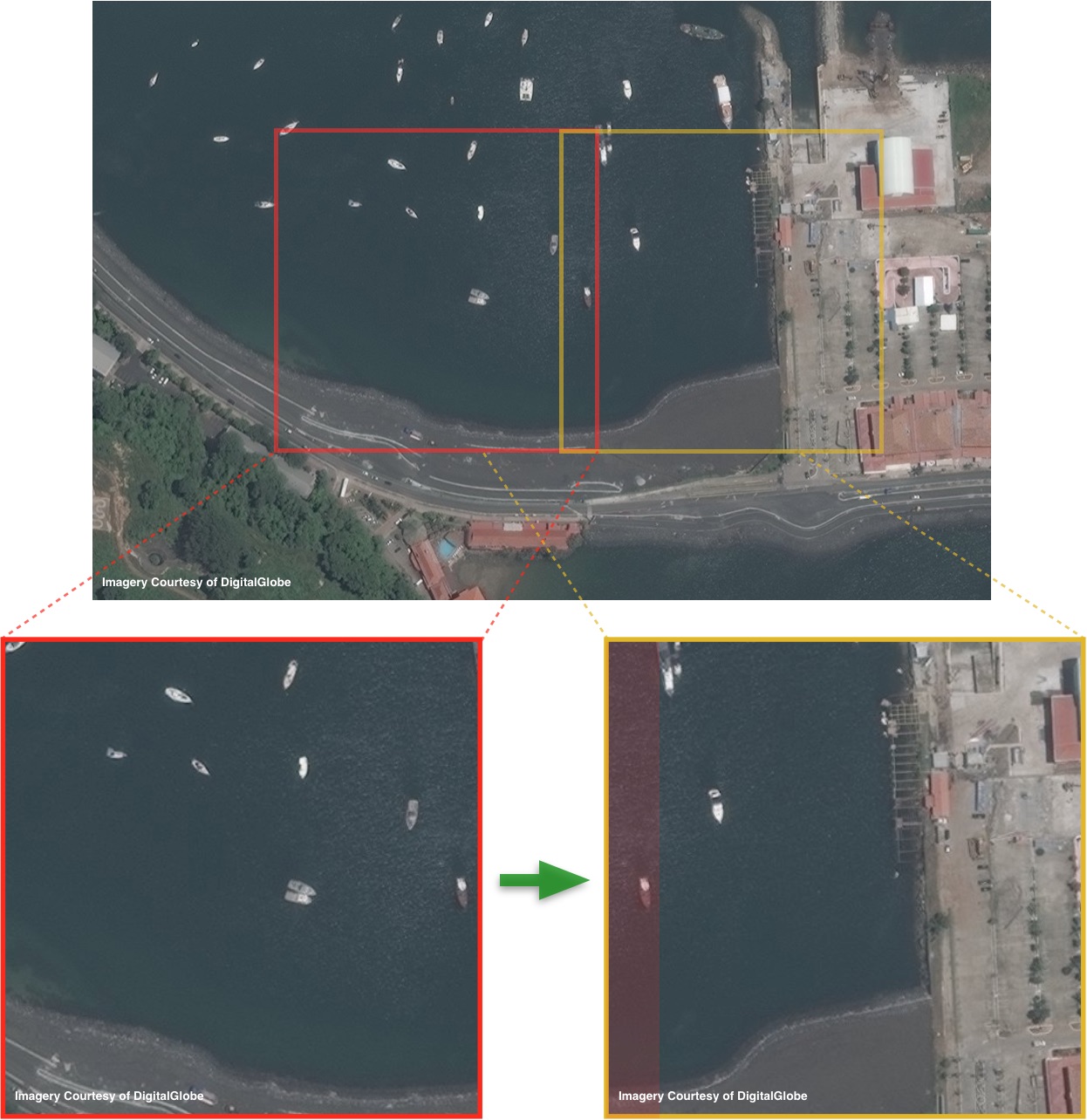}
\end{center}
\caption{Graphic of testing procedure for large image sizes, showing a sliding window going from left to right across Figure \ref{fig:panama_zooma}. The overlap of the bottom right image is shown in red.  Non-maximal suppression of this overlap is necessary to refine detections at the edge of the cutouts.}
\label{fig:slid_win}
\end{figure}

At test time, we 
partition testing images of arbitrary size into manageable cutouts and run each cutout through our trained model.
Partitioning takes place via a sliding window with user defined bin sizes and overlap ($15\%$ by default),
see Figure \ref{fig:slid_win}.  We record the position of each sliding window cutout by naming each cutout according to the schema:

\texttt{ImageName|row\underline{{ }}column\underline{{ }}height\underline{{ }}width.ext}

\noindent
For example:

\texttt{panama50cm|1370\underline{{ }}1180\underline{{ }}416\underline{{ }}416.tif}

\subsection{Post-Processing}\label{sec_post_proc}

Much of the utility of satellite (or aerial) imagery lies in its inherent ability to map large areas of the globe.  Thus, small image chips are far less useful than the large field of view images produced by satellite platforms.  The final step in the object detection pipeline therefore seeks to stitch together the hundreds or thousands of testing chips into one final image strip.  

For each cutout the bounding box position predictions returned from the classifier are adjusted according to the \texttt{row} and \texttt{column} values of that cutout; this provides the global position of each bounding box prediction in the original input image.  The $15\%$ overlap ensures all regions will be analyzed, but also results in overlapping detections on the cutout boundaries.  We apply non-maximal suppression to the global matrix of bounding box predictions to alleviate such overlapping detections.

\section{Training Data}\label{sec_train}


Training data is collected from small chips of large images from three sources: DigitalGlobe satellites, Planet satellites, and aerial platforms. 
Labels are comprised of a bounding box and category identifier for each object.  
We initially focus on five categories: airplanes, boats, building footprints, cars, and airports. 
For objects of very different scales (\eg airplanes vs airports) we show in Section \ref{scale_conf_mit}  that using two different detectors at different scales is very effective.

\begin{figure}[t]
\begin{center}
\includegraphics[width=0.95\linewidth]{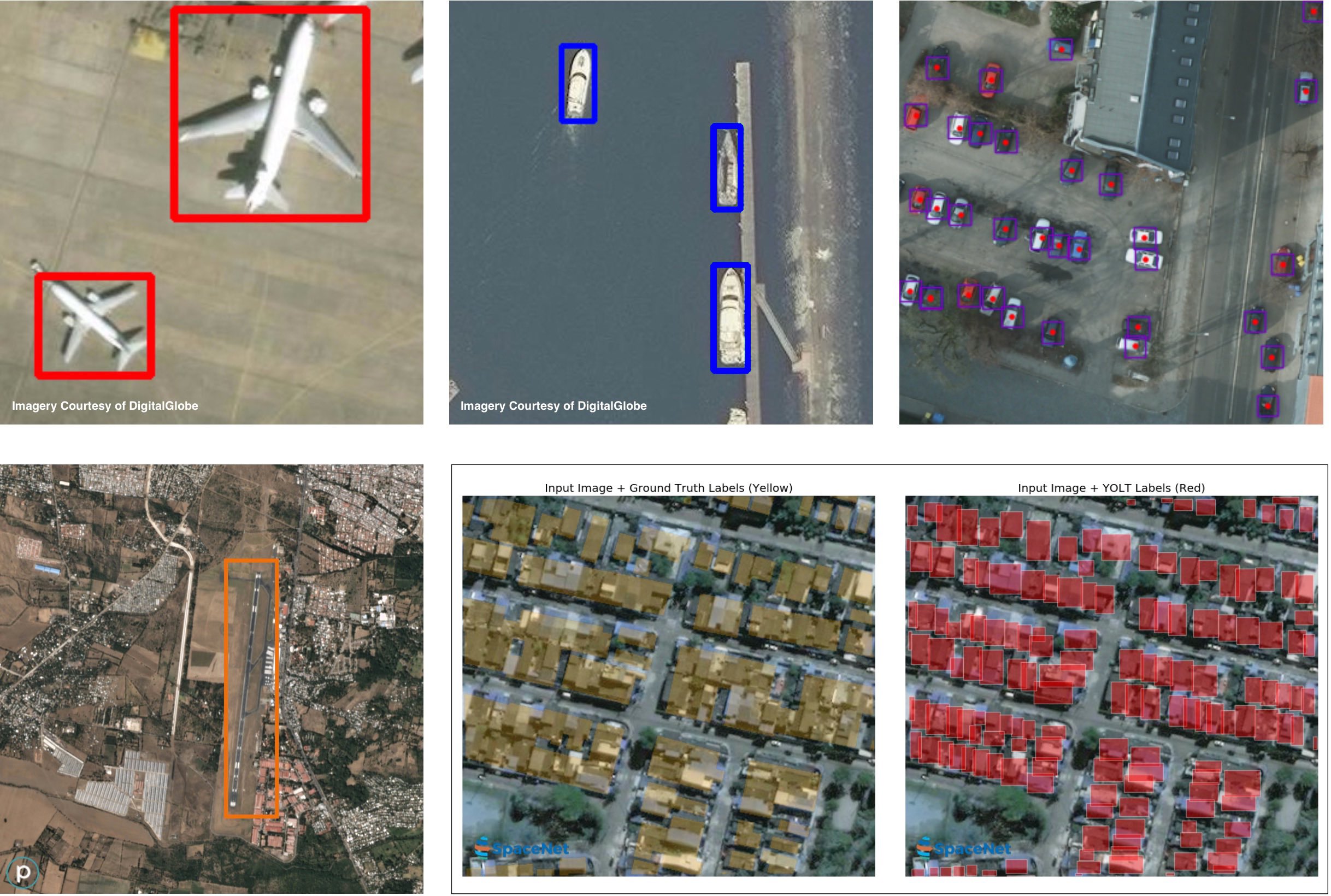}
\end{center}
\caption{YOLT Training data. The top row displays imagery and labels for vehicles.  The top left panel shows airplanes labels overlaid on DigitalGlobe imagery, while the middle panel displays boats overlaid on DigitalGlobe data.  The top right panel shows aerial imagery of cars from the COWC dataset \cite{cowc}, with the red dot denoting the COWC label and the purple box our inferred 3 meter bounding box. The lower left panel shows  an airport (orange) in $4\times$ downsampled Planet imagery.  The lower middle panel shows SpaceNet building footprints in yellow, and the lower right image displays inferred YOLT bounding box labels in red.}
\label{fig:YOLT_training}
\end{figure}

\begin{description}
\item[Cars] The Cars Overhead with Context (COWC) \cite{cowc} dataset is a large, high quality set of annotated cars from overhead imagery collected over multiple locales.  Data is collected via aerial platforms, but at a nadir view angle such that it resembles satellite imagery.
The  imagery has a resolution of 15 cm GSD that is approximately double the current best resolution of commercial satellite imagery (30 cm GSD for DigitalGlobe).  Accordingly, we convolve the raw imagery with a Gaussian kernel and reduce the image dimensions by half to create the equivalent of 30 cm GSD images.  Labels consist of simply a dot at the centroid of each car, and we draw a 3 meter bounding box around each car for training purposes.  We reserve the largest geographic region (Utah) for testing, leaving 13,303 labelled training cars.  

\item[Building Footprints] The second round of SpaceNet data consists of 30 cm GSD DigitalGlobe imagery and labelled building footprints over four cities: Las Vegas, Paris, Shanghai, and Khartoum.  
The labels are precise building footprints, which we transform into bounding boxes encompassing $90\%$ of the extent of the footprint.  Image segmentation approaches show great promise for this challenge; nevertheless, we explore YOLT performance on building outline detection, acknowledging that since YOLT outputs bounding boxes it will never achieve perfect building footprint detection for complex building shapes.  Between the four cities there are 221,336 labelled buildings.

\item[Airplanes] We label eight DigitalGlobe images over airports for a total of 230 objects in the training set.  

\item[Boats] We label three DigitalGlobe images taken over coastal regions for a total of 556 boats.

\item[Airports] We label airports in 37 Planet images for training purposes, each with a single airport per chip. For objects the size of airports, some downsampling is required, as runways can exceed 1000 pixels in length even in low resolution Planet imagery; we therefore downsample Planet imagery by a factor of four for training purposes.

\end{description}

The raw training datasets for airplanes, airports, and watercraft are quite small by computer vision standards, and a larger dataset may improve the inference performance detailed in Section~\ref{sec_results}.



We train with stochastic gradient descent and maintain many of the hyper parameters of \cite{yolo9000}: 5 boxes per grid, an initial learning rate of $10^{-3}$, a weight decay of 0.0005, and a momentum of 0.9.  Training takes $2-3$ days on a single NVIDIA Titan X GPU.

\section{Test Images}

To ensure evaluation robustness, all test images are taken from different geographic regions than training examples.  For cars, we reserve the largest geographic region of Utah for testing, yielding 19,807 test cars. Building footprints are split 75/25 train/test, leaving 73,778 test footprints. We label four airport test images for a total of 74 airplanes.  Four boat images are labelled, yielding 771 test boats.  Our dataset for airports is smaller, with ten Planet images used for testing.  See Table~\ref{tab_train_test_split} for the train/test split for each category.

\begin{table}[t]
\begin{threeparttable}
\centering
\caption{Train/Test Split} 
\label{tab_train_test_split}
\begin{tabular}{l c c}
Object Class	& Training Examples & 	Test Examples  \\
\hline
Airport\tnote{$\ast$}		& 37 &		10 \\
Airplane\tnote{$\ast$}	& 230 &	 	74\\
Boat\tnote{$\ast$}		& 556 &	  	100 \\
\hline
Car\tnote{$\dagger$}		& 19,807 &  	13,303 \\
Building\tnote{$\dagger$}	& 221,336  & 	73,778 \\ 
\end{tabular}
\begin{tablenotes}
\item[$\ast$] Internally Labelled
\item[$\dagger$] External Dataset
\end{tablenotes}
\end{threeparttable}
\end{table}

\section{Object Detection Results}\label{sec_results}

\subsection{Universal Classifier Object Detection Results}

\begin{figure}[t]
\begin{center}
\includegraphics[width=0.95\linewidth]{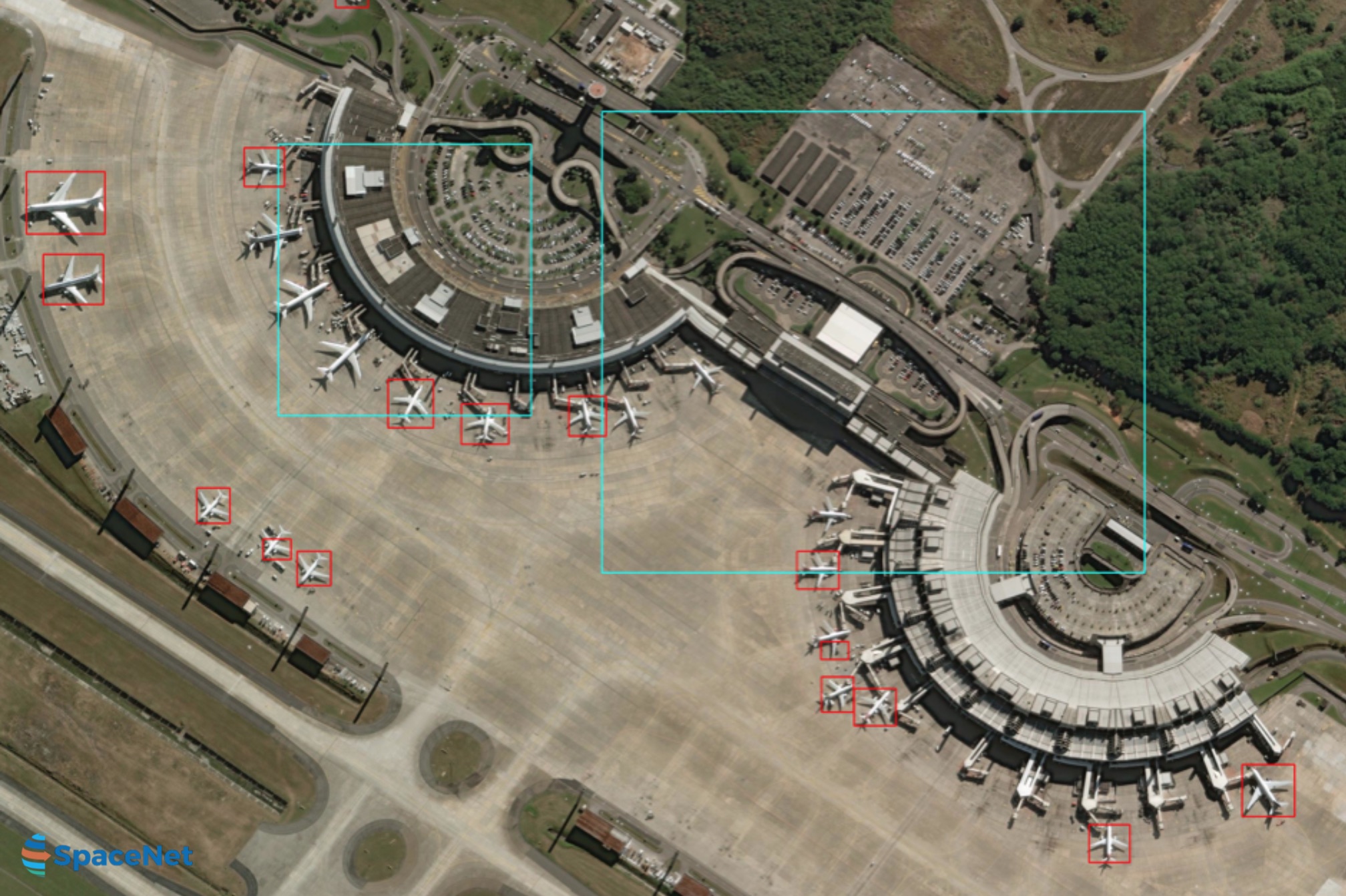}
\end{center}
\caption{Poor results of the universal model applied to DigitalGlobe imagery at two different scales (200m, 1500m). Airplanes are in red. The cyan boxes mark spurious detections of runways, caused in part by confusion from small scale linear structures such as highways.}
\label{fig:airport_spurious}
\end{figure}

Initially, we attempt to train a single classifier to recognize all five categories listed above, both vehicles and infrastructure.  We note a number of spurious airport detections in this example (see Figure \ref{fig:airport_spurious}), as down sampled runways look similar to highways at the wrong scale.  

\subsection{Scale Confusion Mitigation} \label{scale_conf_mit}

There are multiple ways one could address the false positive issues noted in Figure \ref{fig:airport_spurious}.  Recall from Section  \ref{sec_train} that for this exploratory work our training set consists of only a few dozen airports, far smaller than usual for deep learning models.  Increasing this training set size might improve our model, particularly if the background is highly varied.  Another option would be to use post-processing to remove any detections at the incorrect scale (\eg an airport with a size of $\sim50$ meters).  Another option is to simply build dual classifiers, one for each relevant scale. 

We opt to utilize the scale information present in satellite imagery and run two different classifiers: one trained for vehicles + buildings, and the other trained only to look for airports.
Running the second airport classifier on down sampled images has a minimal impact on runtime performance, since in a given image there are approximately 100 times more 200 meter chips than 2000 meter chips.

\subsection{Dual Classifier Results} \label{sec_vehicle}

For large validation images, we run the classifier at two different scales: 200m, and 2500m. The first scale is designed for vehicles and buildings, and the larger scale is optimized for large infrastructure such as airports. We break the validation image into appropriately sized image chips and run each image chip on the appropriate classifier. The myriad results from the many image chips and multiple classifiers are combined into one final image, and overlapping detections are merged via non-maximal suppression.
We find a detection probability threshold of between 0.3 and 0.4 yields the highest F1 score for our validation images. 

We define a true positive as having an intersection over union (IOU) of greater than a given threshold. An IOU of 0.5 is often used as the threshold for a correct detection, though as in Equation 5 of ImageNet \cite{imagenet} we select a lower threshold for vehicles since we are dealing with very small objects.  For SpaceNet building footprints and airports we use an IOU of 0.5. 



\begin{figure}[t]
\begin{center}
\includegraphics[width=0.95\linewidth]{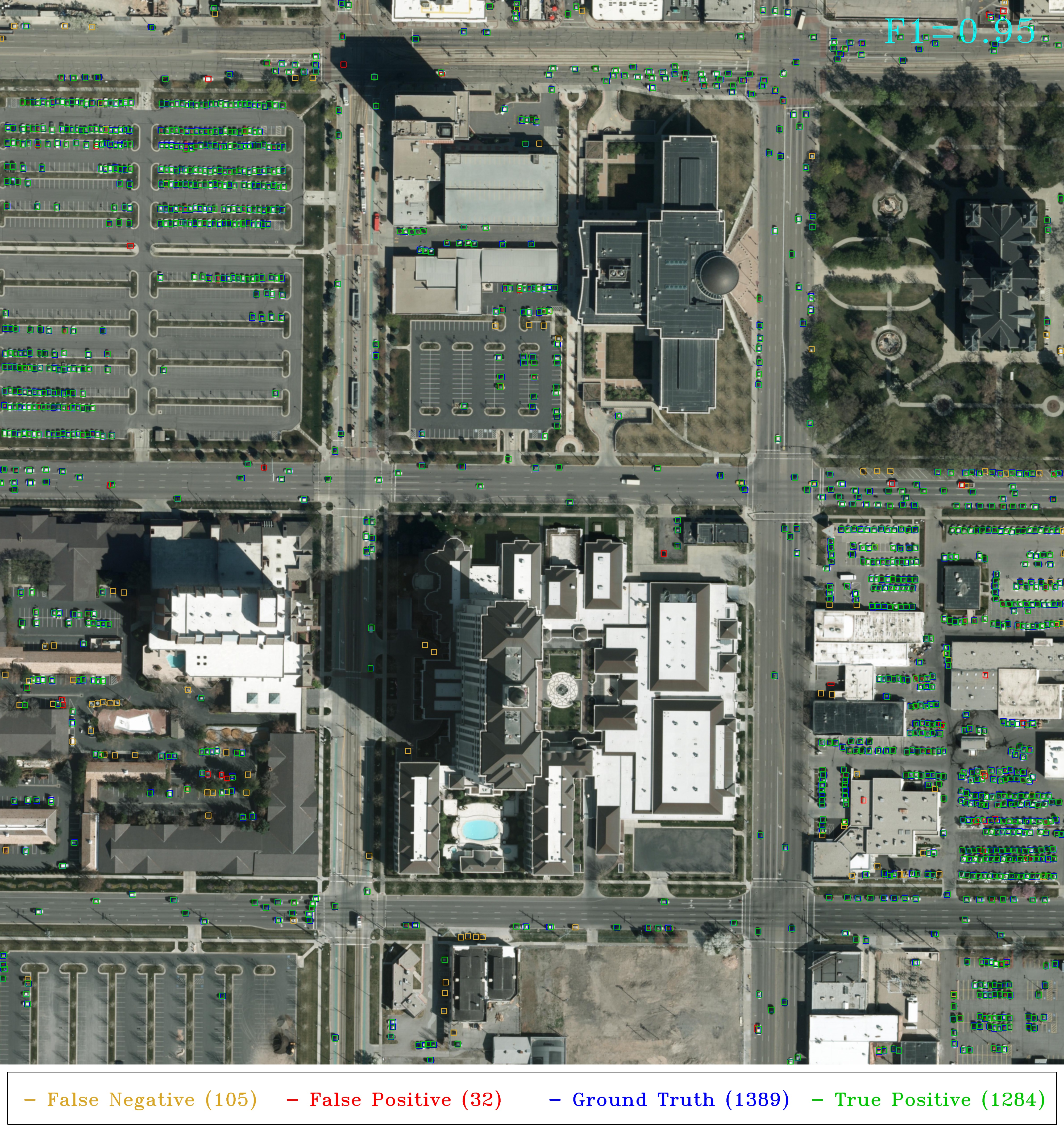} 
\end{center}
\caption{Car detection performance on a $600\times600$ meter aerial image over Salt Lake City ($\rm{Image ID} = 21$) at 30 cm GSD with 1389 cars present.  False positives are shown in red, false negatives are yellow, true positives are green, and blue rectangles denote ground truth for all true positive detections.  
F1 = 0.95 for this test image, and GPU processing time is $<1$ second. 
}
\label{fig:fig_cowc}
\end{figure}

\begin{figure}[t]
\begin{center}
\includegraphics[width=0.95\linewidth]{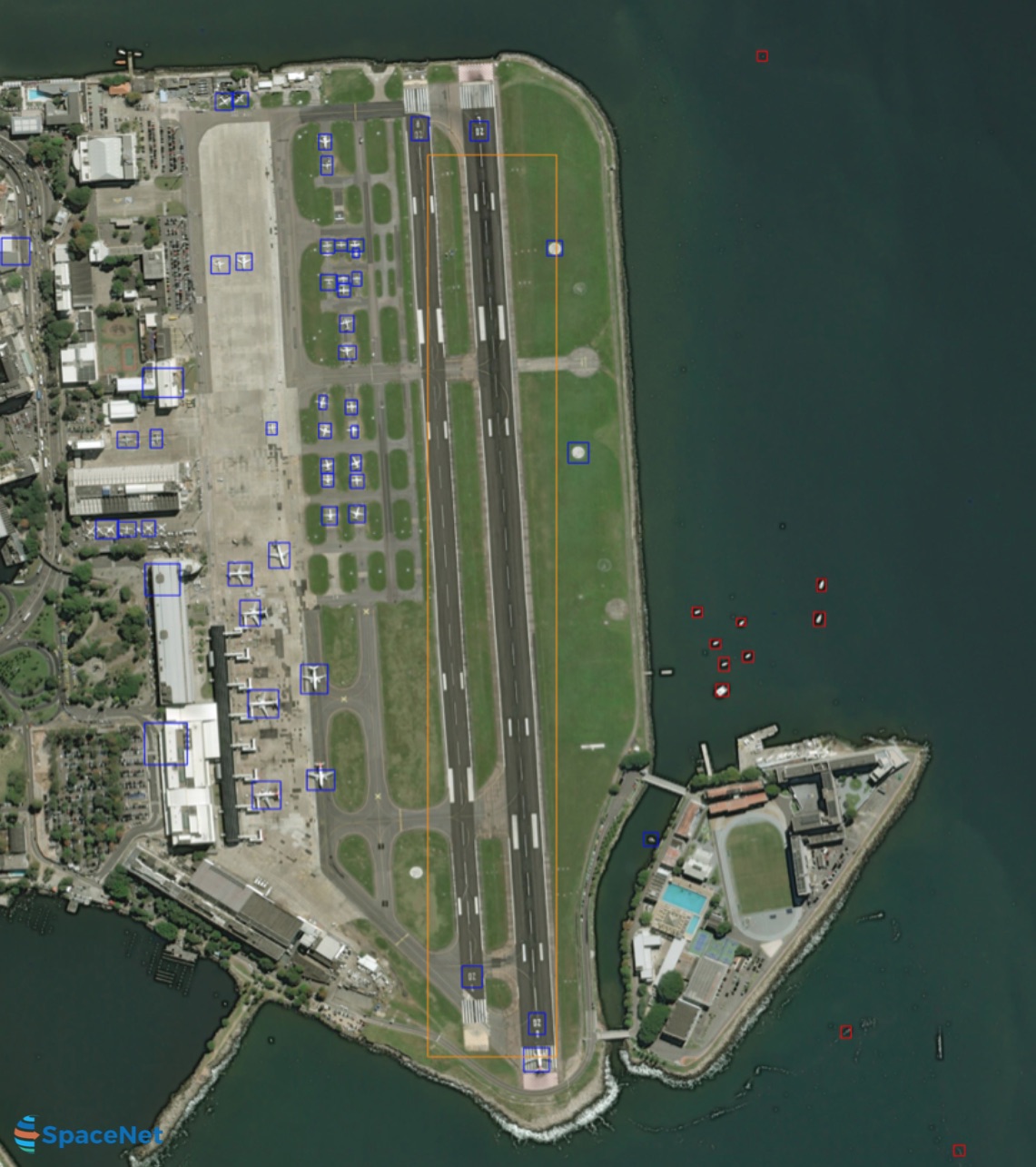}
\end{center}
\caption{YOLT classifier applied to a SpaceNet DigitalGlobe 50 cm GSD image containing airplanes (blue), boats (red), and runways (orange). 
In this image we note the following F1 scores: airplanes = 0.83, boats = 0.84, airports = 1.0.}
\label{fig:fig_comb}
\end{figure}

Table \ref{tab_f1_speed} displays object detection performance and speed over all test images for each object category.  YOLT performs relatively well on airports, airplanes, and boats, despite small training set sizes.  YOLT is not optimized for building footprint extraction, though performs somewhat competitively on the SpaceNet dataset; the top score on the recent SpaceNet challenge achieved an F1 score of 0.69\footnote{https://spacenetchallenge.github.io/Competitions/Competition2.html}, while the YOLT score of 0.61 puts it in the top 3.    
We report inference speed in terms of GPU time to run the inference step.  
Inference runs rapidly on the GPU, at $\sim 50$ frames per second.  
Currently, pre-processing (\ie splitting test images into smaller cutouts) and post-processing (\ie stitching results back into one global image) is not fully optimized and is performed on the CPU, which adds a factor of $\approx2$ to run time.
The inference speed translates to a runtime of $<6$ minutes to localize all vehicles in an area of the size of Washington DC, and $<2$ seconds to localize airports over this area.  DigitalGlobe's WorldView3 
satellite\footnote{http://worldview3.digitalglobe.com} 
covers a maximum of 680,000 km$^2$ per day, so at YOLT inference speed a 16 GPU cluster would provide real-time inference on satellite imagery.

\begin{table}[t]
\begin{threeparttable}
\centering
\caption{YOLT Performance and Speed} 
\label{tab_f1_speed}
\begin{tabular}{l c c}
Object Class	& F1 Score & 	Run Time  \\
& &  ($\rm{km^2/min}$) \\
\hline

Car\tnote{$\dagger$}		& $0.90 \pm 0.09$ & 32 \\
Airplane\tnote{$\ast$}	& $0.87 \pm 0.08$ &	 32\\
Boat	\tnote{$\ast$}		& $0.82 \pm 0.07$ &	32 \\
Building\tnote{$\ast$}  	& $0.61 \pm 0.15$  & 32 \\ 
Airport\tnote{$\ast$}		& $0.91 \pm 0.14$ &	6000 \\

\end{tabular}
\begin{tablenotes}
\item[$\dagger$] IOU = 0.25
\item[$\ast$] IOU = 0.5
\end{tablenotes}
\end{threeparttable}
\end{table}

\subsection{Detailed Performance Analysis}\label{sec_res}

The large test set of $\sim20,000$ cars in the nine Utah images of the COWC dataset enables detailed performance analyses.  The majority of the cars ($> 13,000$) lie in the image over central Salt Lake City so we split this image into sixteen smaller $600\times600$ meter regions to even out the number of cars per image. 
We remove one test scene that has only 61 cars in the scene, leaving 23 test scenes, with mean count per test image of  $1130 \pm 540$.  We apply a YOLT model trained to find cars on these test scenes.

\begin{figure}[t]
\begin{center}
\includegraphics[width=0.95\linewidth]{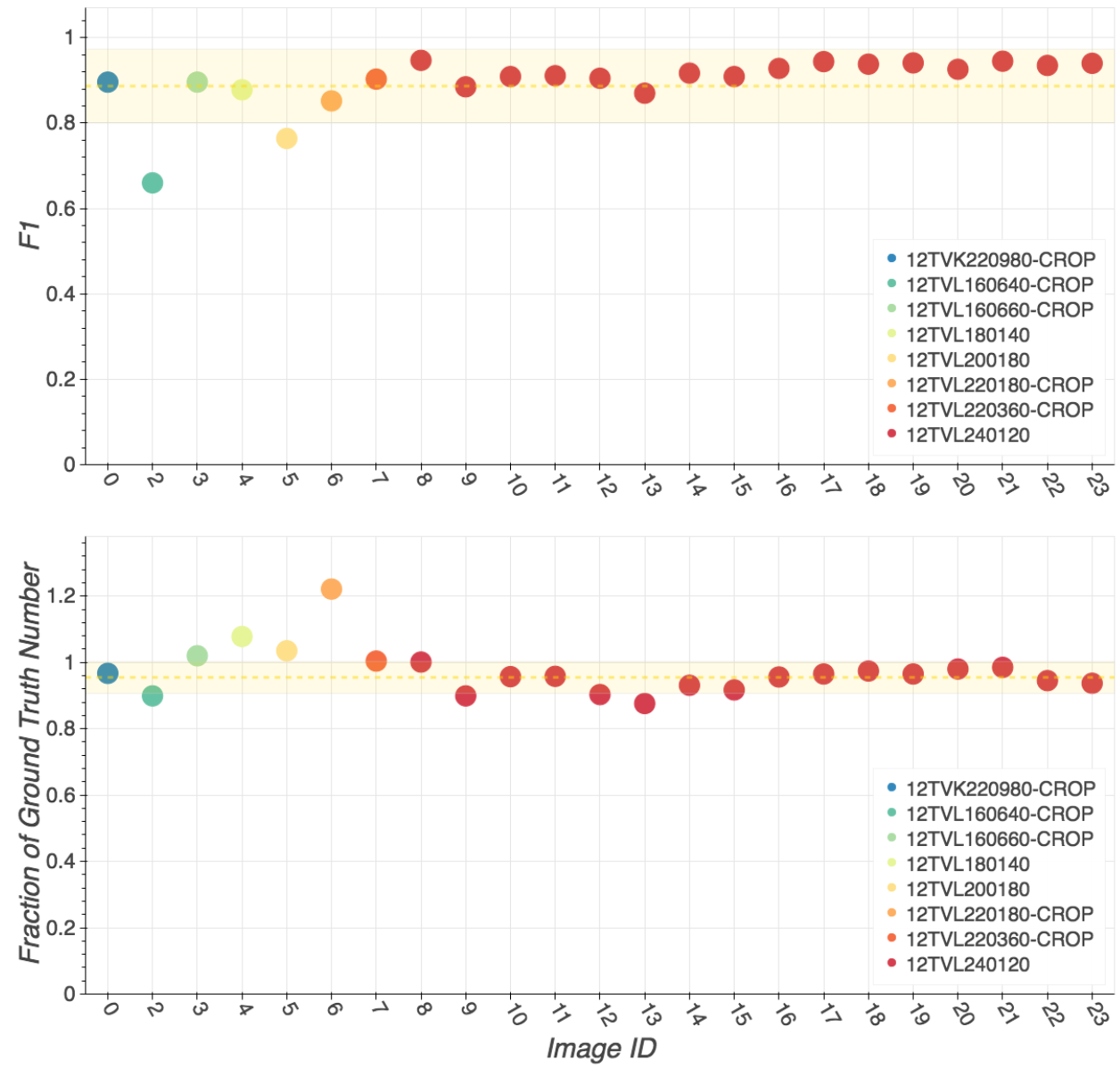}
\end{center}
\caption{Top: F1 score per COWC test scene.  ($F1 = 0.90 \pm 0.09$).  Bottom: Number of detections as a fraction of ground truth number ($F_c = 0.95 \pm 0.05$.  Dot colors correspond to the test scene, with the multiple red dots indicating central Salt Lake City cutouts. The dotted orange line denotes the weighted mean, with the yellow band displaying the weighted standard deviation.}
\label{fig:cowc_f1+frac}
\end{figure}

In Figure \ref{fig:cowc_f1+frac} we display the F1 score for each scene, along with the car count accuracy.  
Total car count in a specified region may be a more valuable metric in the commercial realm than F1 score. Accordingly, we compute the number of predicted cars for each scene as a fraction of ground truth number ($F_c = N_{\rm{predicted}} / N_{\rm{truth}})$.
Like the F1 score, a value of 1.0 denotes perfect prediction for the fractional car count metric. The COWC \cite{cowc} authors sought to count (rather than localize) the number of cars in test images, and achieved an error of $0.19\%$.  Total count error for YOLT on the COWC data is $0.90\%$.


Inspection of Figure \ref{fig:cowc_f1+frac} reveals that the F1 score and ground truth fraction are quite high for typical urban scenes, (\eg 
$\rm{Image ID} = 21$ shown in  Figure \ref{fig:fig_cowc}).  The worst outlier in Figure \ref{fig:cowc_f1+frac} is $\rm{Image ID} = 2$, with an F1 score of 0.67, and 2860 cars present.  This location corresponds to an automobile junkyard, an understandably difficult region to analyze.

\section{Resolution Performance Study}

The uniformity of object sizes in the COWC \cite{cowc} dataset enables a detailed resolution study.  To study the effects of resolution on object detection, we convolve the raw 15 cm imagery with a Gaussian kernel and reduce the image dimensions to create additional training and testing corpora at [0.30, 0.45, 0.60, 0.75, 0.90, 1.05, 1.20, 1.50, 1.80, 2.10, 2.40, 3.00] meters.  

\begin{figure}[t]
\begin{center}
\includegraphics[width=0.95\linewidth]{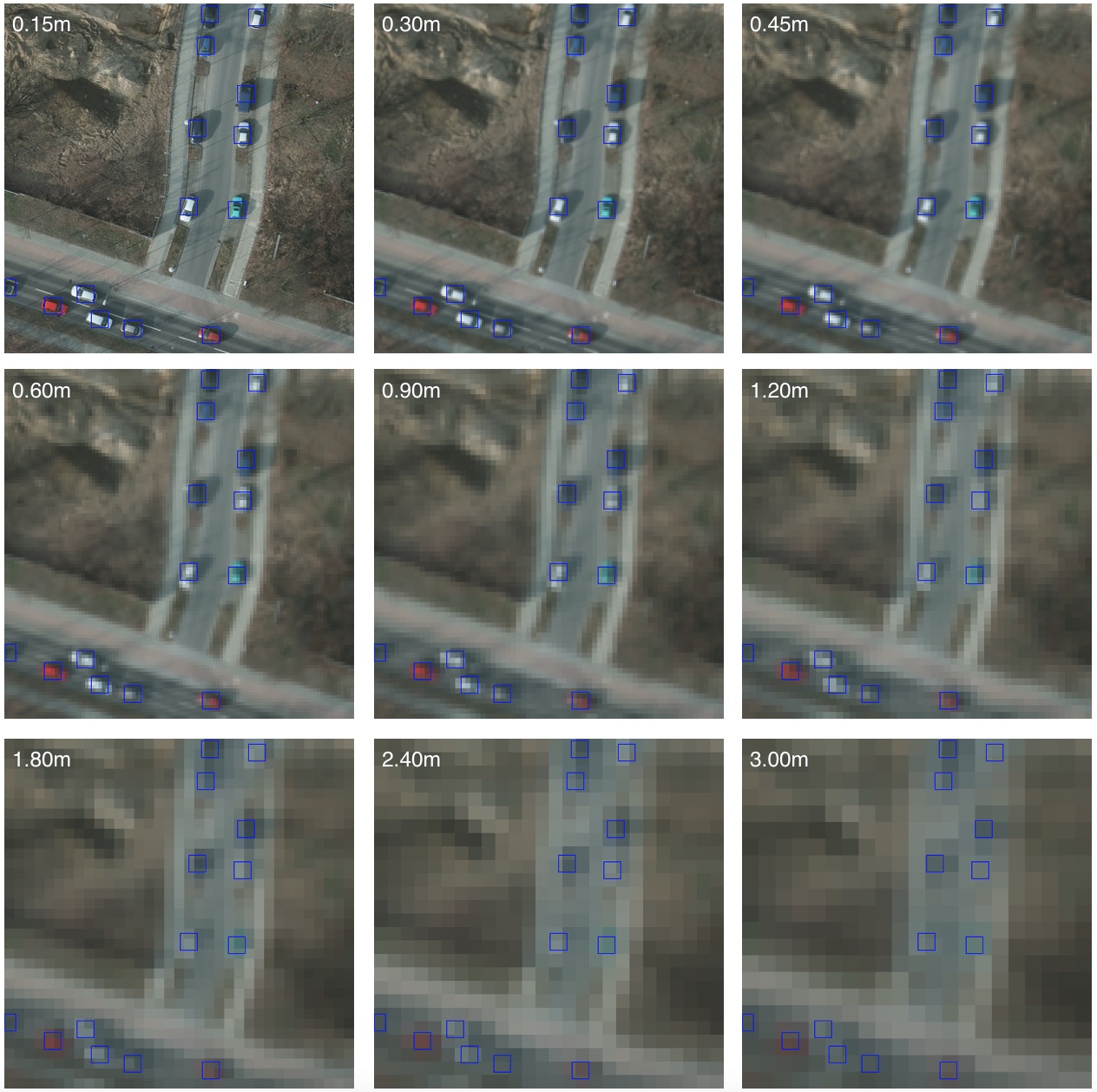}
\end{center}
\caption{COWC \cite{cowc} training data convolved and resized to various resolutions from the original 15 cm resolution (top left); bounding box labels are plotted in blue. 
}
\label{fig:fig_cowc_grid}
\end{figure}

\begin{figure}[t]
\begin{center}
\includegraphics[width=0.95\linewidth]{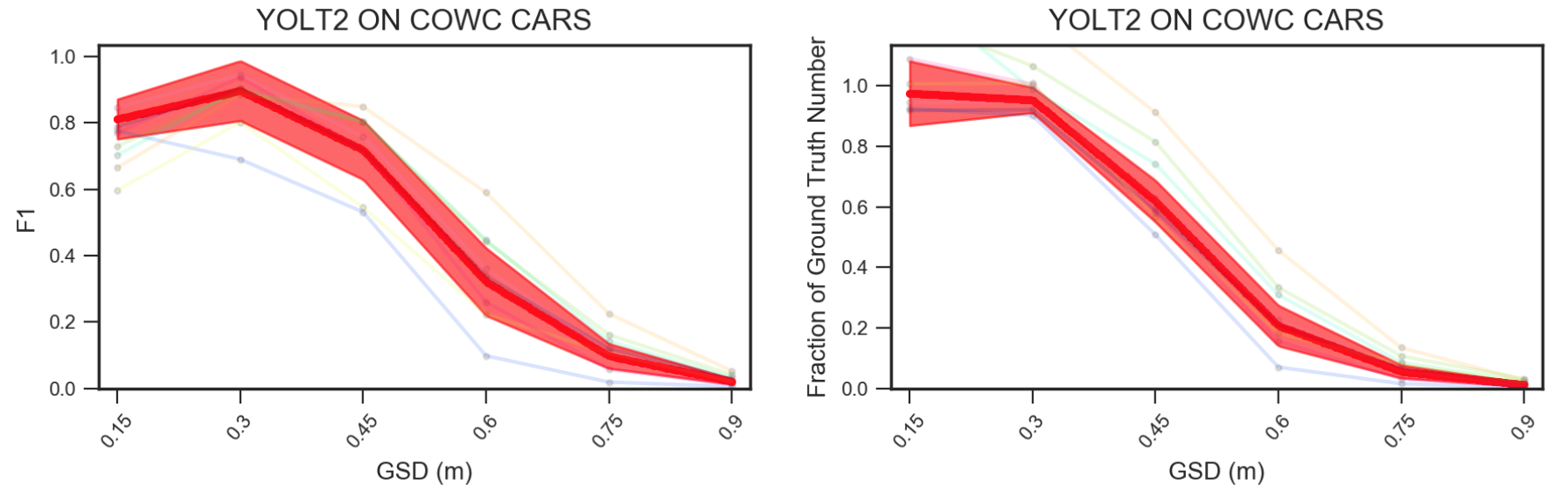}
\end{center}
\caption{Performance of the $0.3$m model applied to various resolutions. The 23 thin lines display the performance of each individual test scene; most of these lines are tightly clustered about the mean, denoted by the solid red. The red band displays $\pm 1$ STD. The model peaks at F1 $= 0.9$ for the trained resolution of 0.3m, and rapidly degrades when evaluated with lower resolution data; it also degrades somewhat for higher resolution 0.15m data. 
}
\label{fig:fig_cowc_comp_score}
\end{figure}

Initially, we test the multi-resolution test data on a single model (trained at 0.30 meters), and in Figure \ref{fig:fig_cowc_comp_score} demonstrate that the ability of this model to extrapolate to multiple resolutions is poor.  
Subsequently, we train a separate model for each resolution, for thirteen models total.  Creating a high quality labelled dataset at low resolution (2.4m GSD, for example) is only possible because we downsample from already labelled high resolution 15 cm data; typically low resolution data is very difficult to label with high accuracy.    

\begin{figure}[t]
\begin{center}
\includegraphics[width=0.95\linewidth]{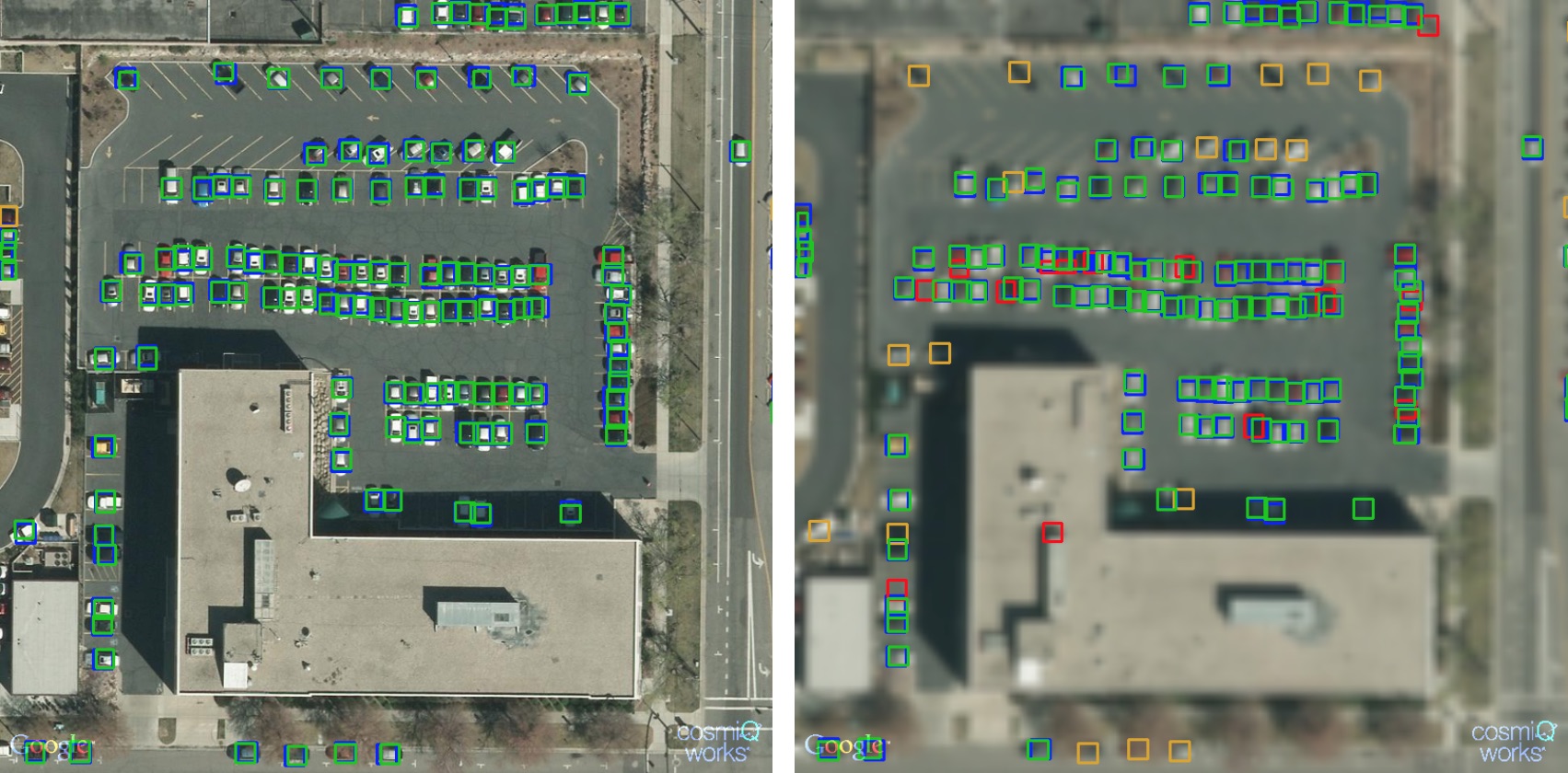}
\end{center} 
\caption{Object detection results on different resolutions on the same $120 \times 120$ meter Salt Lake City cutout of COWC data.
The cutout on the left is at 15 cm GSD, with an F1 score of 0.94, while the cutout on the right is at 90 cm GSD, with an F1 score of 0.84.  
 }
\label{fig:fig_cowc_dual}
\end{figure}

\begin{figure}[t]
\begin{center}
\includegraphics[width=0.95\linewidth]{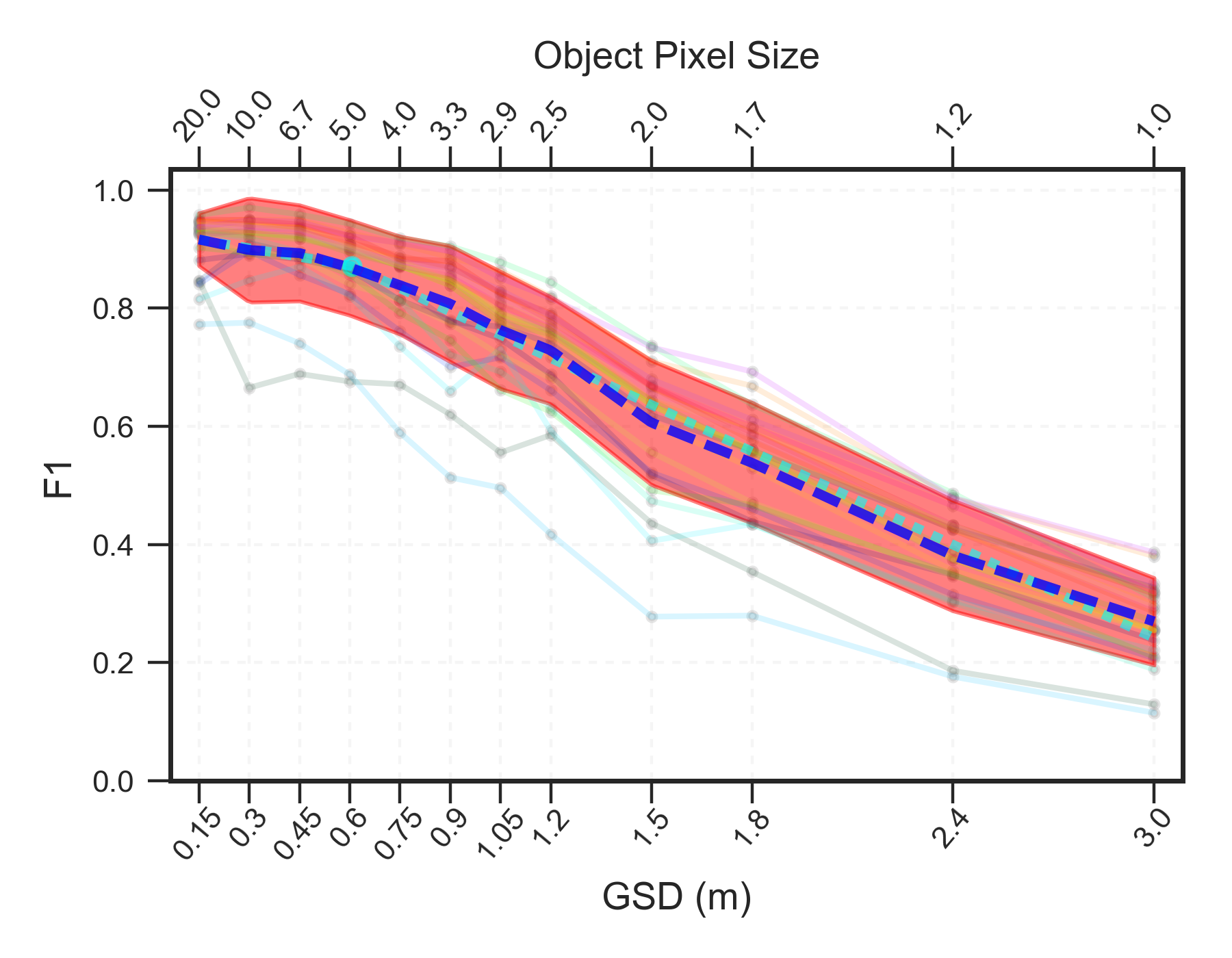}
\end{center}
\caption{Object detection F1 score for ground sample distances of $0.15 - 3.0$ meters (bottom axis), corresponding to car size  of  $20 - 1$ pixel(s) (top axis).  At each of the thirteen resolutions we evaluate test scenes with a unique model trained at that resolution.  The 23 thin lines display the performance of the individual test scenes; most of these lines are tightly clustered about the mean, denoted by the blue dashed line. The red band displays $\pm 1$ STD.  We fit a piecewise linear model to the data, shown as the dotted cyan line. Below the inflection point (large cyan dot) of 0.61 meters (corresponding to a car size of 5 pixels) the F1 score degrades slowly with a slope of $\Delta F1/\Delta GSD = -0.10$; between 0.60 m and 3.0 m GSD the slope is steeper at $-0.26$.  The F1 scores at 0.15 m, 0.60 m, and 3.0 m GSD are 0.92, 0.87, and 0.27, respectively. }
\label{fig:fig_cowc_f1_score}
\end{figure}

\begin{figure}[t]
\begin{center}
\includegraphics[width=0.95\linewidth]{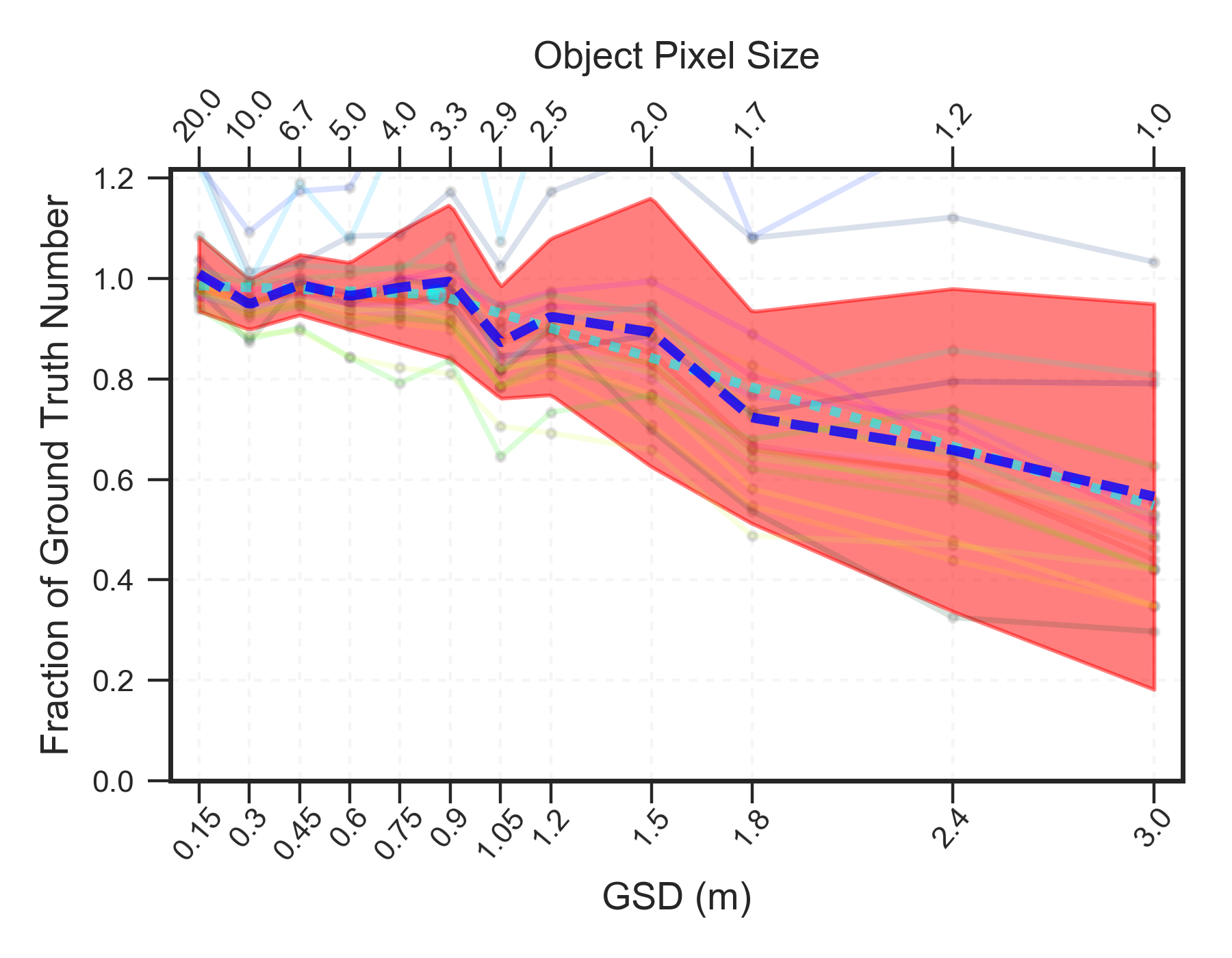}
\end{center}
\caption{Fraction of predicted number of cars to ground truth, with a unique model for each resolution (bottom axis) and object pixel size (top axis). A fraction of 1.0 means that the correct number of cars was predicted, while if the fraction is below 1.0 too few cars were predicted. The thin bands denote the performance of the 23 individual scenes, with the dashed blue line showing the weighted mean and the red band displaying $\pm 1$ STD. We fit a piecewise linear model to the data, shown as the dotted cyan line. Below the inflection point (large cyan dot) of $0.86$ meters the slope is essentially flat with a slope of $-0.03$; between $0.87$ m and $3$ m GSD the slope is steeper at $-0.20$. For resolutions sharper than $0.86$ meters the predicted number of cars is within $4\%$ of ground truth. }
\label{fig:fig_cowc_num_score}
\end{figure}

For objects $\sim3$ meters in size we observe from Figure \ref{fig:fig_cowc_f1_score} that object detection performance degrades from $F1=0.92$ for objects 20 pixels in size to $F1=0.27$ for objects 1 pixel in size, with a mean error of 0.09. Interestingly, the F1 score only degrades by only $\approx5\%$ as objects shrink from 20 to 5 pixels in size (0.15m to 0.60m GSD). At least for cars viewed from overhead, one can conclude that object sizes of $\geq 5$ pixels yield object detection scores of $F1 > 0.85$.  
The curves of Figure \ref{fig:fig_cowc_comp_score}  degrade far faster than Figures \ref{fig:fig_cowc_f1_score} and \ref{fig:fig_cowc_num_score}, illustrating that a single model fit at high resolution is inferior to a series of models trained at each respective resolution.

\section{Conclusions}
Object detection algorithms have made great progress as of late in localizing objects in 
ImageNet style datasets.  Such algorithms are rarely well suited to the object sizes or orientations present in satellite imagery, however, nor are they designed to handle images with hundreds of megapixels.  

To address these limitations we implemented a fully convolutional neural network pipeline (YOLT) to rapidly localize vehicles, buildings, and airports in satellite imagery.  We noted poor results from a combined classifier due to confusion between small and large features, such as highways and runways.  Training dual classifiers at different scales (one for buildings/vehicles, and one for infrastructure), yielded far better results.

This pipeline yields an object detection F1 score of $\approx 0.6 - 0.9$, depending on category.  While the F1 scores may not be at the level many readers are accustomed to from ImageNet competitions, object detection in satellite imagery is still a relatively nascent field and has unique challenges.  In addition, our training dataset for most categories is relatively small for supervised learning methods, and
the F1 scores could possibly be improved with further post-processing of detections.  

We also demonstrated the ability to train on one sensor (\eg DigitalGlobe), and apply our model to a different sensor (\eg Planet).   We show that at least for cars viewed from overhead, object sizes of $\geq 5$ pixels yield object detection scores of $F1 > 0.85$. 
The detection pipeline is able to evaluate satellite and aerial images of arbitrary input size at native resolution, and processes vehicles and buildings at a rate of $\approx 30 \, \rm{km}^2$ per minute, and airports at a rate of $\approx 6,000 \, \rm{km}^2$ per minute.  At this inference speed, a 16 GPU cluster could provide real-time inference on the DigitalGlobe WorldView3 satellite feed.  
\newline


\begin{acks}
  We thank Karl Ni for very helpful comments.  
\end{acks}

\bibliographystyle{ACM-Reference-Format}
\bibliography{yolt_bib}

\end{document}